\documentclass[10pt,journal,compsoc]{IEEEtran}
\ifCLASSOPTIONcompsoc
  \usepackage[nocompress]{cite}
\else
  \usepackage{cite}
\fi
\ifCLASSINFOpdf
\else
\fi
\usepackage{cite}
\usepackage{amsmath,amssymb,amsfonts}
\usepackage{algorithmic}
\usepackage{textcomp}
\usepackage{multicol}
\usepackage{lipsum}
\usepackage{longtable}
\usepackage{vcell}
\usepackage{wrapfig}
\usepackage{algorithm}
\usepackage{epsfig}
\usepackage{graphicx}
\usepackage{booktabs}
\usepackage{multirow}
\usepackage{hhline}
\usepackage{textcomp}
\usepackage{url}
\usepackage[leftcaption]{sidecap}
\usepackage{caption}
\usepackage{xcolor, soul}
\usepackage{colortbl}
\usepackage{caption}
\usepackage{subfig}

\begin{document}
%
\title{Neural Networks for Infectious Diseases Detection: Prospects and Challenges}
%
%
%
%

\author{Muhammad Azeem,
        Shumaila Javaid,
        Hamza Fahim and Nasir~Saeed,~\IEEEmembership{Senior~Member,~IEEE }
\IEEEcompsocitemizethanks{\IEEEcompsocthanksitem M. Azeem is with University of Sargodha, Pakistan.
.\protect\\
E-mail: azeemchaudharyg@gmail.com
\IEEEcompsocthanksitem S. Javaid and H. Fahim are with Tongji University, China.
\protect\\
Email: shumaila@tongji.edu.cn, hamzafahim@tongji.edu.cn
\IEEEcompsocthanksitem N. Saeed is with the Department
of Electrical, King Abdullah University of Science and Technology, Thuwal, Makkah, Saudi Arabia
.\protect\\
E-mail: mr.nasir.saeed@ieee.org

}
}

\IEEEtitleabstractindextext{%
\begin{abstract}
Artificial neural network (ANN) ability to learn, correct errors, and transform a large amount of raw data into useful medical decisions for treatment and care have increased its popularity for enhanced patient safety and quality of care. Therefore, this paper reviews the critical role of ANNs in providing valuable insights for patients' healthcare decisions and efficient disease diagnosis. We thoroughly review different types of ANNs presented in the existing literature that advanced ANNs adaptation for complex applications. Moreover, we also investigate ANN's advances for various disease diagnoses and treatments such as viral, skin, cancer, and COVID-19. Furthermore, we propose a novel deep Convolutional Neural Network (CNN) model called ConXNet for improving the detection accuracy of COVID-19 disease. ConXNet is trained and tested using different datasets, and it achieves more than 97\% detection accuracy and precision, which is significantly better than existing models. Finally, we highlight future research directions and challenges such as complexity of the algorithms, insufficient available data, privacy and security, and integration of biosensing with ANNs. These research directions require considerable attention for improving the scope of ANNs for medical diagnostic and treatment applications.
\end{abstract}

\begin{IEEEkeywords}
Artificial neural networks,  Healthcare, Convolutional Neural Networks, Infectious diseases.
\end{IEEEkeywords}}

\maketitle

\section{Introduction}
\label{sec:introduction}

Artificial intelligence (AI) is revolutionizing the world with its endless applications. The platforms built on AI  are omnipresent, accelerating the pace of developing life-saving drugs and reducing operations costs. The building block of AI technologies is the neural networks, called artificial neural networks (ANN), that stimulate the human brain's analyzing and processing abilities to solve complex problems. The unique characteristics of ANN (such as efficient data handling, low complexity, reduced computation and storage requirements) have huge potential for a wide range of disciplines, including medical sciences \cite{gil2009application} (especially in the areas of cardiology \cite{vseckanovic2020review}, radiology \cite{yamashita2018convolutional}, oncology \cite{djavan2002novel}, urology \cite{checcucci2020applications}), veterinary \cite{lisboa2002review}, stock exchange \cite{bahrammirzaee2010comparative}, law enforcement department \cite{edwards2015systematic}, ecology \cite{brosse1999predicting}, human resource management \cite{vellido1999neural}, and cybersecurity \cite{saleem2019evaluating}. The ANN are mainly based on mathematical models inspired by biological nervous systems such as brain routes information. The ANN works similarly to an adaptive system, which updates its configuration in the learning phase and can be modeled for a specific application, such as data classification and pattern categorization. A neural network consists of three basic layers (i.e., input layer, hidden layer,  and output layer). Based on these layers, ANN can be categorized into single-layer Feed-Forward Neural Network (FFNN) \cite{scarselli1998universal}, multilayer feed-forward network \cite{abiodun2018state}, a single node with its feedback, and a multilayer recurrent network \cite{de2015survey}.

The potential of these different types of ANN models in the healthcare industry cannot be ignored. For instance, the studies in \cite{doyle2000infectious}  have extensively explored the potential of ANN for diagnosis and treatment of various infectious diseases, including Tuberculosis \cite{miller2000population}, diarrhea \cite{hennessy2004survey}, dengue \cite{tseng2009estimating}, childhood blindness \cite{zhang2018development}, and COVID-19  \cite{wieczorek2020neural, Saeed2020COVID}.  Most of the time, accuracy and complexity are the major performance metrics to evaluate these ANN. For instance, Agrawal et al. in \cite{agrawal2015neural} studied the efficiency of ANN for cancer detection. They highlighted that most neural networks indicate great results in accurately classifying tumor cells, especially Multi-Layer Perceptron (MLP) and Probabilistic Neural Network (PNN) provide the highest accuracy of 97.1\% and 96\%, respectively.
 The comprehensive study of neural networks for cancer detection also reveals that neural network techniques deliver a good classification rate; however, their training time is excessively high \cite{ganesan2010application}
In another work in \cite{yadav2016skin}, ANN and image processing techniques-based diagnosis systems are studied for skin diseases. Furthermore, \cite{dagadkhair2021survey}  presented a survey on widely spread parasite infectious disease (commonly known as Malaria infection) detection using CNNs \cite{dagadkhair2021survey}.  Similarly, reference \cite{muhammad2020deep} reviews the literature on using deep learning for brain tumor classification, focusing on preprocessing, features extraction, and classification for brain tumor identification. Moreover, \cite{azad2021medical} discusses various ML algorithms such as Naïve Bayes, Classification and Regression Tree (CART), Decision Tree (DT), and Support Vector Machine (SVM) for different diseases such as lungs cancer, breast cancer, and skin diseases.  Consequently, in \cite{zand2015comparative}, the authors present a comparative study on data mining techniques, including Naïve Bayes, back-propagated neural network, and decision tree algorithms for breast cancer prediction.

In \cite{kaur2017survey}, Kaur et al. investigate the effectiveness of numerous nature-inspired computing techniques such as genetic algorithms \cite{tomassini1995survey}, ant colony optimization \cite{dorigo2005ant}, particle swarm optimization \cite{zhang2015comprehensive}, and artificial bee colony \cite{karaboga2014comprehensive} for diagnosing various critical human disorders. They conclude that nature-inspired computing techniques have high accuracy for disease detection and diagnosis. However, the survey lacks a detailed comparison with state-of-the-art schemes required to investigate the efficiency of the nature-inspired techniques for real-world problems. Another study in \cite{philemon2019review} extensively discusses hybrid models (i.e., neural networks combined with other classical methods or meta-heuristics approaches) to outline selecting well-suited ANN model for the epidemic forecasts. Reference \cite{philemon2019review} shows that hybrid neural networks depict enhanced performance for epidemic forecasts.

Besides other infectious diseases detection,  recently, ANN are explored for COVID-19 detection. For instance, \cite{chen2020survey} reviews a broad view of AI technology for diagnosing and treating COVID-19. Reference \cite{chen2020survey} highlights that disease detection and diagnosis, virological research, drug and vaccine development, epidemic and transmission prediction, medical image analysis, and drug discovery are the major areas that integrate AI to fight against COVID-19.  Moreover, the authors in \cite{ulhaq2020COVID} present the
role of computer vision in combating the COVID-19 pandemic. They consider three types of visionary images for COVID-19 detection: Computed Tomography (CT) scans, X-Ray Imagery, and Ultrasound imaging.

Unlike the existing surveys, our paper briefly reviews the progression of ANN over time for diseases diagnosis and discusses their applicability for major diseases, such as skin, cancer, and COVID-19 detection. We have also critically studied the state-of-the-art CNN models for COVID-19 detection to understand their impact on the detection and treatment of COVID-19. In addition, we proposed ConXNeT, a novel framework that significantly enhances the COVID-19 detection. The key contributions of our presented review are summarized as follows:

\begin{itemize}
    \item First, we review state-of-the-art ANN models and critically summarizes their part in advancing detection and diagnosis of various diseases such as Alzheimer, skin, cancer and retinal, and COVID-19.
    \item Then, we present the studies on the existing CNN models for COVID-19 detection and highlights their contributions and limitations. Moreover, in contrast to existing surveys, our paper also proposes ConXNet, a novel deep learning model, trained and tested using different datasets to improve the detection accuracy of COVID-19 up to 98\%.
    \item Finally, we highlight the gaps that require attention in the future to improve the ANN-based disease diagnosis and treatment. These future research challenges include complexity of the algorithms, insufficient available data, privacy and security, and integration of biosensing with ANNs.
\end{itemize}

The rest of the paper is organized as follows: Section \ref{background} thoroughly discusses the background of ANN. Section \ref{covidmodels} provides a comprehensive study of existing CNN models designed for COVID-19.  In Section \ref{ConXNet}, we provide the complete demonstration of our proposed ConXNet model, including the details of datasets and obtained results for COVID-19 detection. Section \ref{diseases} reviews existing CNN models for various diseases  and Section \ref{directions} highlight the challenges and future research directions. Finally, Section \ref{conclusion} summarizes the findings of the presented reviews and provides future work directions.

\section{Background of Artfical Neural Networks (ANNs)} \label{background}
Warren McCulloch, who was n American neurophysiologist, first described ANN in 1943 that led to the development of neural networks with electrical circuits by Walter Pitts \cite{mcculloch1943logical}. Later in 1949, Donald Hebb also supported the idea of neurons in his book \cite{hebb1955drives} and pointed out that neural trails are reinforced each time they are used. Following the growing interest in ANN, Nathanial Rochester from the IBM research laboratories simulated the first neural network in the 1950s \cite{mccarthy2006proposal}. In 1960, Frank Rosenblatt, a neuro-biologist of Cornell, invented the first perceptron inspired by the neural working of a human brain by doing some simulations  \cite{rosenblatt1960perceptron}. Then, Frank Rosenblatt invented the perceptron at the Cornell Aeronautical Laboratory to cognize human memory, learning, and cognitive processes. He revealed the Mark I Perceptron, the first machine that could "learn" to distinguish and identify optical patterns \cite{estebon1997perceptrons}. Since the invention of the first perceptron in the 1960s, a very long gap has occurred in the advancement of neural networks. However, in 1982, John Hopfield re-motivated the scientific community by presenting a new perspective of building devices using neural networks.


In recent times, neural networks have gained particular importance due to their diverse applications. The research and scientific communities are optimistic regarding the potentials of ANN.  The core property of an ANN is its capability of learning. There are three types of learning: 1) Supervised learning which is accomplished in the existence of an observer. In this type of learning, a supervisor or observer is compulsory for error minimization.  2) Unsupervised learning, which is accomplished without the help of an observer, and the network itself discover features, categories, patterns, or symmetries from the input data and associations for the input data over the output. 3) Reinforcement learning, which is used to train the machine learning models. It enables an agent to learn through the consequences of actions in a specific environment.

Over the years, many neural networks are developed and categorized based on their aforementioned learning characteristics. In the following, we extensively discuss different types of neural networks to comprehend the background of ANN.

\subsection{Perceptron}
The perceptron model, proposed by Minsky-Papert in the 1950s, is one of the modest and ancient models of the neuron. Perceptron is the smallest component of a neural network that performs particular calculations to identify features in the input data \cite{park1996predicting}.  Perceptron is a supervised learning algorithm that can classify the data into two categories by a hyperplane; therefore, it is also called a binary classifier. The perceptron algorithm has gained significant attention recently due to its use in establishing logic gates such as AND, OR, or NAND \cite{ali2017forecasting}. Nevertheless, perceptron can only identify linearly separable problems like Boolean AND operation. In contrast, it does not work for non-linear hitches such as the Boolean XOR problem because the classes in XOR are not linearly separable. A straight line cannot be drawn to separate the points in case of a non-linear data \cite{epley1966reviews}.

\begin{figure}
    \centering
    \includegraphics[width=0.5\textwidth]{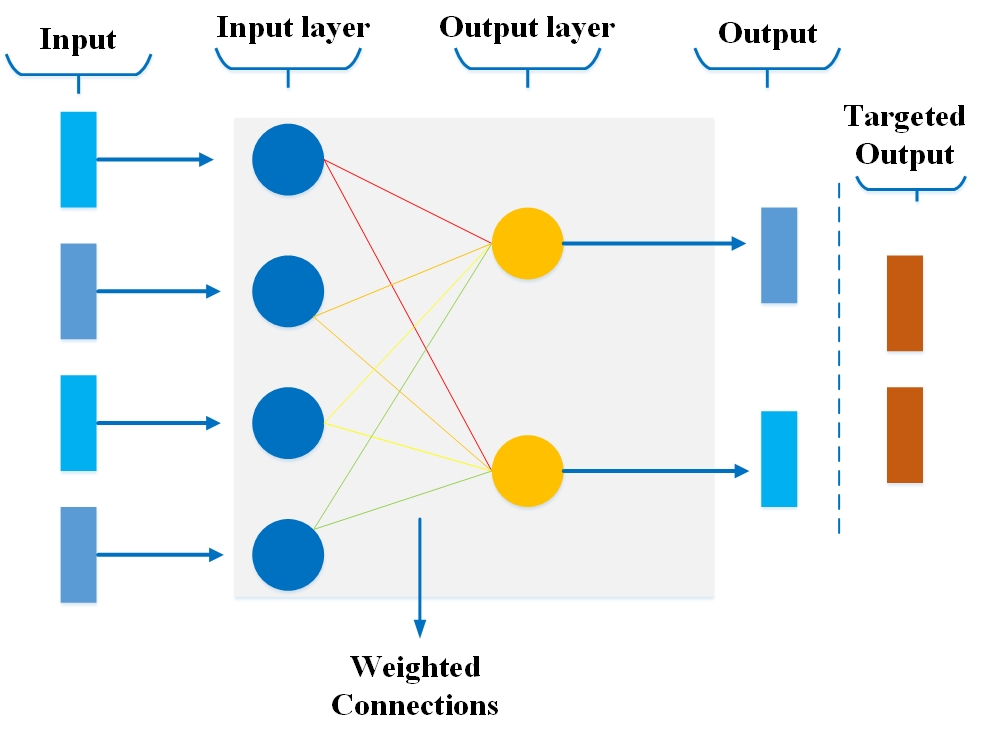}
    \caption{Perceptron architecture}
    \label{fig:pre}
\end{figure}

Fig. \ref{fig:pre} shows a simple perceptron as an example with a layer of input nodes (layer 1) and a layer of output nodes (layer 2). Every perceptron layer is fully connected, but no links occur among nodes in the same layer. When layer 1 directs a signal to layer 2, the related weights on the links are applied, and each acceptance node on layer 2 sums up the received values. If the sum surpasses a given threshold, that node, in turn, directs an output signal \cite{ali2017forecasting}.

The outputs are summed through all the inputs ($a[i]$) received by a node ($j$) in the output layer. The output of each node as
\begin{equation}\label{eq:1}
    S_j=\sum_{i=0}^{n}a_iw_{ij}
\end{equation}
if $S_j > \theta$ then  $x_j=1$, otherwise $x_j=0$, where $\theta$ is the threshold. The perceptron’s output can be "trained" to match the desired output by adjusting the weights on the connections between layers. The amount of the correction is determined by multiplying the difference between the actual output ($x[j]$) and target ($t[j]$) values by a learning rate constant ($C$). If the nodes’ output ($a[i]$) is 1, that connection weight is adjusted, and if it sends 0, it has no bearing on the output, and subsequently, there is no need for adjustment. Thus, the process wight adjustment is as follows
\begin{equation}
    w^{ij}_{new}= w^{ij}_{old} + C(t_j - x_j)a_i,
\end{equation}
where C is the learning rate. This training procedure is repeated until the network performance reaches a maximum threshold \cite{ali2017forecasting}.

\subsection{Multilayer Perceptron}
\begin{figure}
    \centering
    \includegraphics[width=0.5\textwidth]{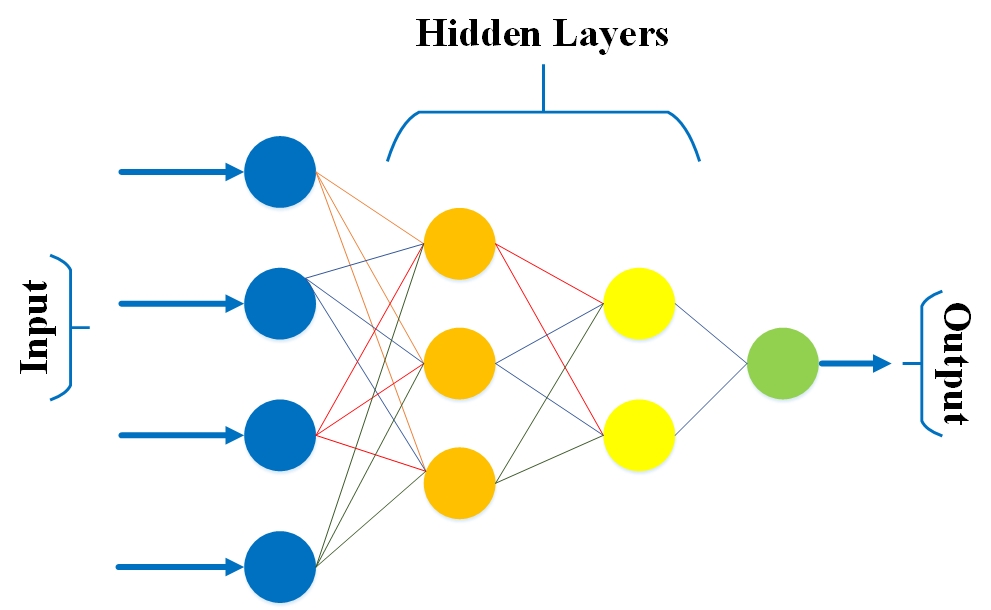}
    \caption{A simple presentation of MLP}
    \label{fig:MLP}
\end{figure}

Multilayer Perceptron (MLP) was first introduced in the 1980s and works as an entry point in the direction of complex neural networks where input data travels over several layers of artificial neurons \cite{gardner1998artificial}. In MLP, every node is associated with all neurons in the subsequent layer, making it a fully connected neural network. Input and output layers exist with many hidden layers (i.e., at least three or more layers in total) with a bi-directional propagation (i.e., forward and backward). In forward propagation, inputs are multiplied with weights and sent to the activation function, and in backpropagation, the weights are adjusted to reduce the loss \cite{ruck1990feature}. Weights in MLP are machine-learned values from neural networks that are self-adjusted according to the difference between predicted outputs versus training inputs. In MLP, nonlinear activation functions are followed by softmax as an output layer activation function. Although MLP is comparatively complex to design, maintain, and slow (depends on the number of hidden layers) \cite{pal1992multilayer}, it has several applications, such as speech recognition, machine translation, and complex classification. Fig. \ref{fig:MLP} shows a simple presentation of MLP, consisting of an input layer, a hidden layer and an output layer. Except for the input nodes, each node is a neuron that uses a nonlinear activation function.

\subsection{Feed Forward Neural Network (FFNN)}
\begin{figure}
    \centering
    \includegraphics[width=0.5\textwidth]{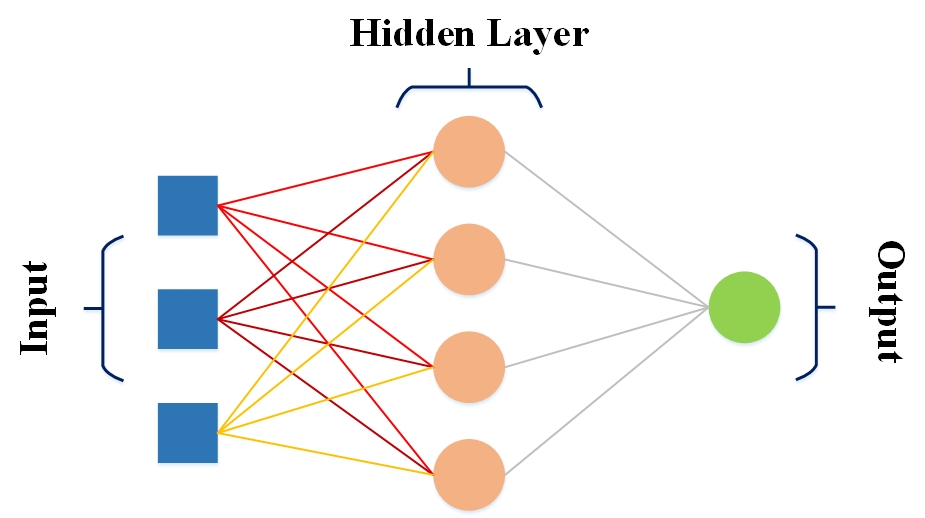}
    \caption{FFNN architecture}
    \label{fig:FFNN}
\end{figure}

FFNN \cite{abiodun2018state} is the most basic neural network used for different applications, including computer vision, natural language processing, and speech recognition. In FFNN, data propagates only in one direction, passing through neural nodes and exiting through output nodes. FFNN comprises input and output layers; however, hidden layers may or may not exist.  The weights are static in FFNN, and an activation function is used by inputs multiplied by weights to learn the value of parameter $\theta$. In addition, the design of FFNN requires considering several crucial factors, including an optimizer, cost function, and output units. The cost function is minimized using an optimizer or optimization algorithm such as Adam \cite{wang2021generative}, stochastic gradient descent \cite{netrapalli2019stochastic}, and Adagrad \cite{sun2019survey}, which adjusts the values of the weights and biases after each training cycle or epoch until the cost function approaches the global optimum. In comparison, output units are the part of the output layer to provide the required output or prediction, thus completing the task that the neural network must complete. FFNN in Fig. \ref{fig:FFNN} shows input passed through one or more hidden layers in a uni-directional way, and weights are multiplied by inputs to minimize the error using an optimizer.

\subsection{Convolutional Neural Network (CNN)}
\begin{figure}
    \centering
    \includegraphics[width=0.3\textwidth]{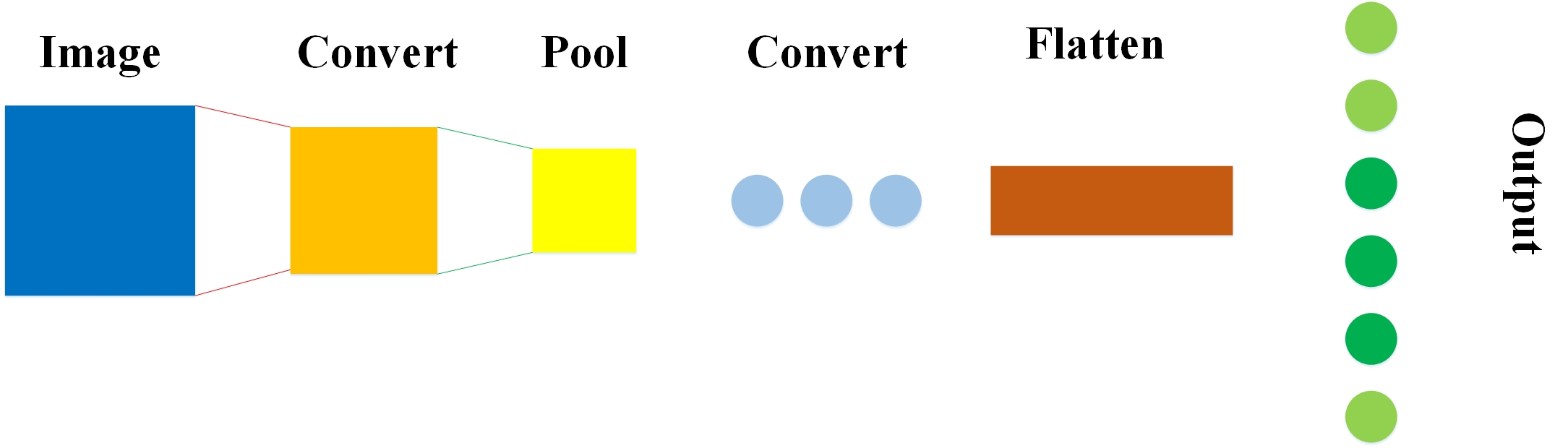}
    \caption{A simple presentation of CNN}
    \label{fig:RCNN}
\end{figure}

CNN was first introduced in the 1990s by Yann LeCun, which could identify handwritten digits \cite{lecun1998gradient}. Initially, the use of CNN was only restricted to image processing, requiring large datasets. However, in 2012 after the success of AlexNet, which win the ImageNet challenge, showed that the time has come to revisit CNNs, as large datasets are available in the shape of ImageNet \cite{alom2018history}. A CNN comprises a three-dimensional set of neurons instead of the typical two-dimensional array. The first layer is called a convolutional layer, where every neuron only routes the information from a minor part of the layer. The convolution layer uses ReLU as an activation function followed by softmax. The first layer is followed by a pooling layer and where the output of the convolution layer drives to a fully connected neural network for classification. The CNNs show very promising results in the fields of image and video recognition, semantic parsing, and paraphrase detection. Fig. \ref{fig:RCNN} visualizes a CNN model that first extracts all the features of the input image using the convolution operation and then uses the pooling layer to extract the most prominent features to pass through to the other convolutional layer for linearity. Later,  the flatten layer shapes the input data for producing output.

\subsection{Radial Basis Function Neural Networks}
\begin{figure}
    \centering
    \includegraphics[width=0.5\textwidth]{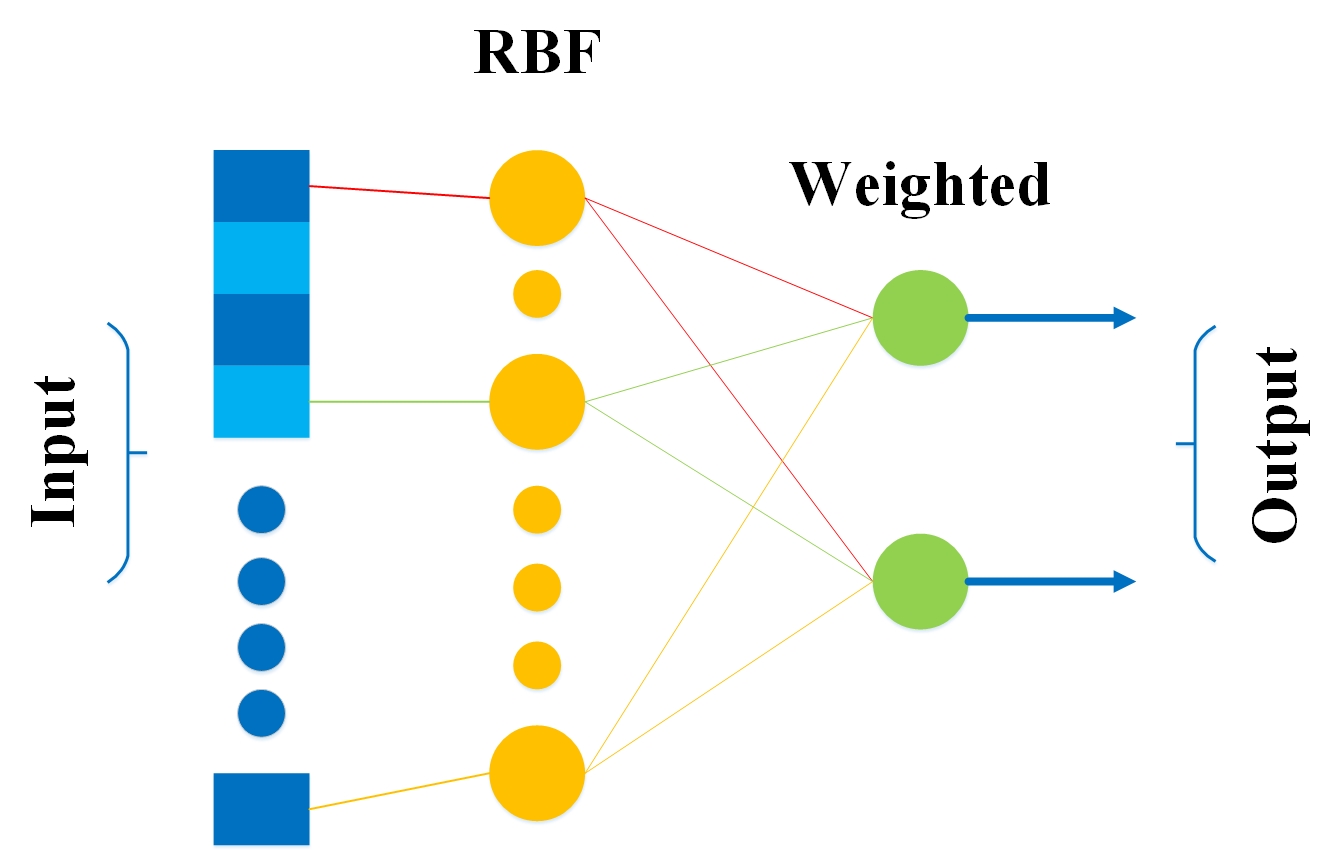}
    \caption{Architecture of RBF}
    \label{fig:RBF}
\end{figure}

Radial Basis Function (RBF) Network comprises an input vector followed by a layer of RBF neurons and an output layer with one node per class. Classification is accomplished by assessing the input’s resemblance to data points from the training set where each neuron stores a prototype, which is one of the samples from the training set. When a novel input vector (the n-dimensional vector that you are trying to classify) needs to be classified, every neuron computes the Euclidean distance between the input and its prototype \cite{bishop1991improving,yingwei1997sequential,er2002face}. For example, if we have two classes, i.e., class A and Class B, then if the new input is much closer to class A, it will be labeled as class A. Each RBF neuron associates the input vector to its prototype and outputs a value alternating, a measure of resemblance from 0 to 1. As the input equals the prototype, the output of that RBF neuron will be 1, and with the distance produced between the input and prototype, the reply falls off exponentially towards 0. The curve produced by the neuron’s response inclines to a typical bell curve \cite{orr1996introduction, park1991universal}. Fig. \ref{fig:RBF} shows the simple architecture of RBF comprising of an input layer, a hidden layer (consisting of several RBF non-linear activation units), and an output layer.

\subsection{Recurrent Neural Networks (RNN)}
\begin{figure}
    \centering
    \includegraphics[width=0.5\textwidth]{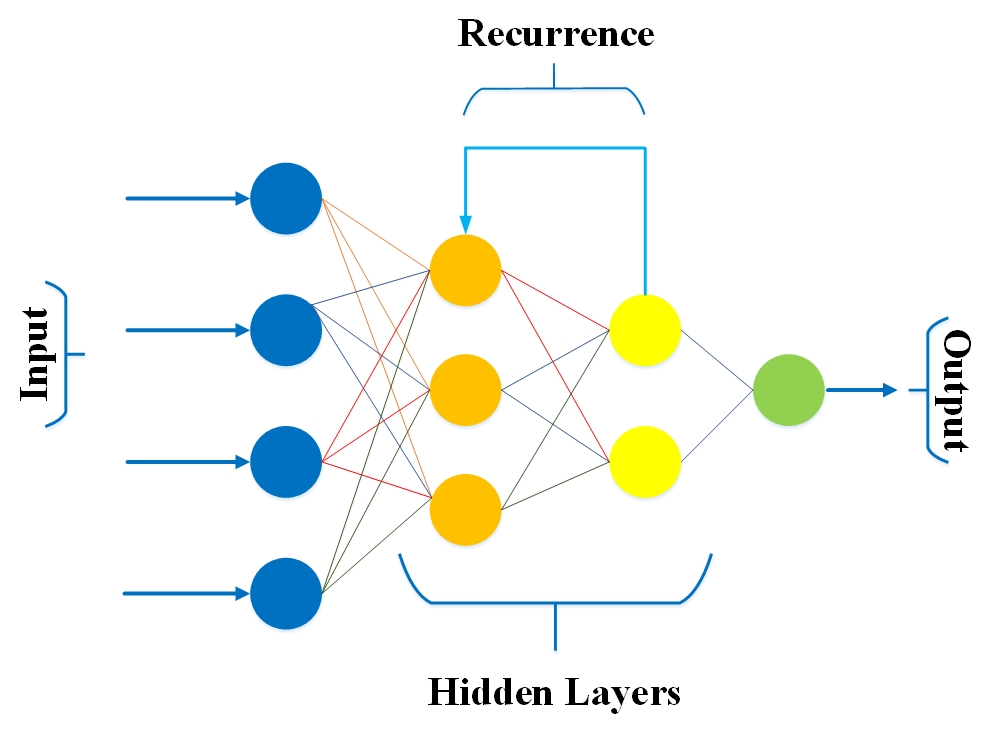}
    \caption{Architecture of RNN}
    \label{fig:RNN}
\end{figure}

RRN is one of the most powerful and robust state-of-the-art algorithms derived from FFNN for modeling sequence data \cite{liu2014efficient}. RNN differs from other algorithms due to its internal memory and ability to process the sequence data that lacks in other algorithms. Furthermore, RNN trains the model based on the current input and previous learning experience, offering more precise predictions. Therefore, RNNs have promising applications in speech recognition, text summarization,  prediction problems, face detection, music composition, and language processing \cite{zimmermann2011historical}. Fig. \ref{fig:RNN} shows a simple RNN architecture consisting of an input layer, one or more hidden layers, and an output layer.

\subsection{Long Short-Term Memory (LSTM) Networks}
\begin{figure}
    \centering
    \includegraphics[width=0.5\textwidth]{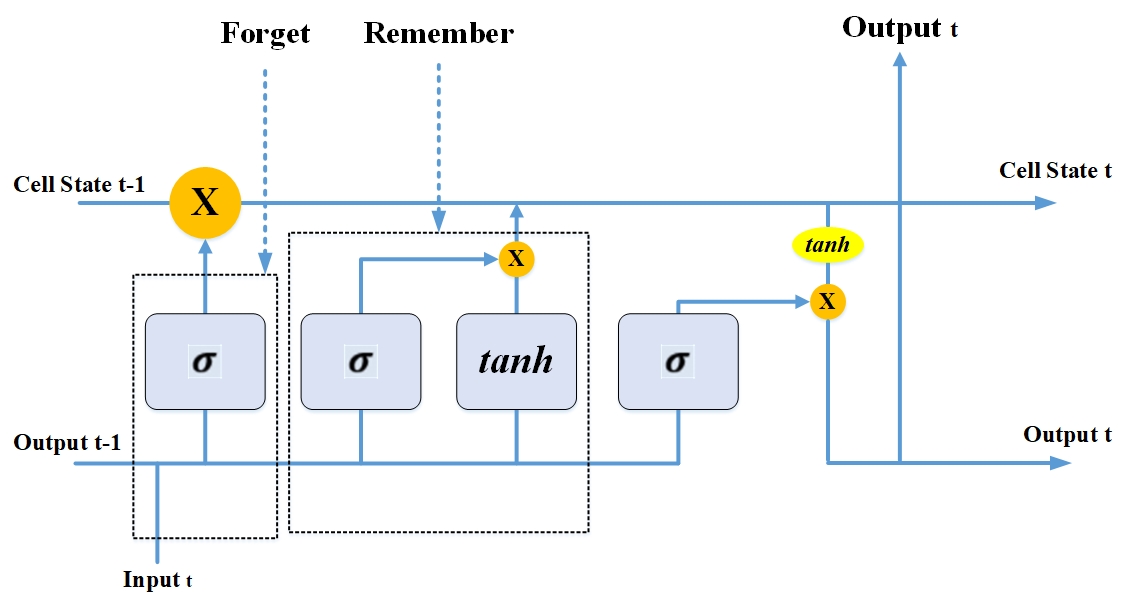}
    \caption{Architecture of LSTM}
    \label{fig:LSTM}
\end{figure}

LSTM architecture was introduced in 1997, which is a type of RNN that uses special units in adding to standard units \cite{sundermeyer2012lstm}. LSTM units comprise a ‘memory cell’ that can preserve information in memory for long periods. A set of gates is used to regulate when information arrives in the memory of a cell. There are three different types of gates, i.e., input gate, output gate, and forget gate. The input gate selects how much information from the last example will be reserved in memory; the output gate controls the amount of data delivered to the next layer, and forget gates regulate the tearing rate of memory keeping. This planning lets them learn longer-term reliance \cite{xingjian2015convolutional}. Fig. \ref{fig:LSTM} show the architecture of LSTM. Fig. \ref{fig:LSTM} shows the architecture of LSTM consisting of three gates i.e., forget gate, remember gate, and output gate, deciding about the data sequence. In LSTM, the \textit{tanh} activation function controls the values that flow across the network whereas the sigmoid activation function $(\sigma)$ is used at the gates. In LSTM, the tanh activation function controls the values that flow across the network whereas the sigmoid activation function $(\sigma)$ is used at the gates as a gating function, it produces an output value of either 0 or 1 that allows either no flow or complete flow of information throughout the gates. In addition, X denotes the input that flows throughout the network to acquire output related to that input.

\subsection{Sequence to Sequence Models}
\begin{figure}
    \centering
    \includegraphics[width=0.5\textwidth]{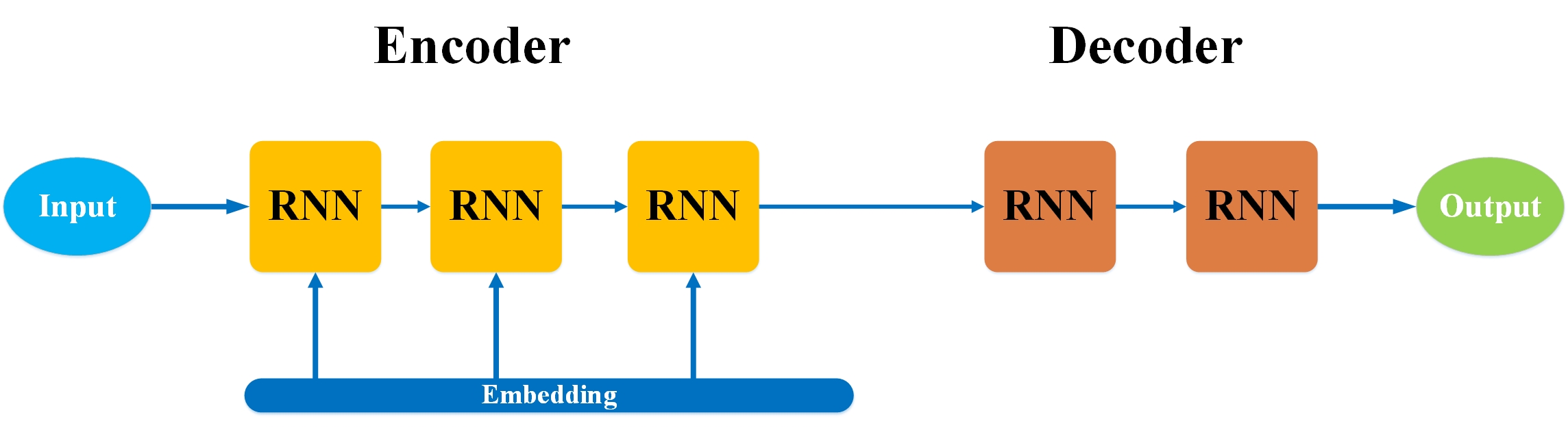}
    \caption{Architecture of sequence to sequence model}
    \label{fig:sequence}
\end{figure}

A sequence to sequence model comprises of two RNNs with additional encoder and decoder modules. The encoder routes the input data, whereas the decoder routes the output. The encoder and decoder work concurrently, either using the same parameters or different ones \cite{chiu2018state}. This model, conflicting with the actual RNN, is primarily applicable in those cases where the length of the input data is equal to the length of the output data \cite{prabhavalkar2017comparison}. Though they own similar assistance and curbs to the RNN, these models are mostly applied in chatbots, machine translations, and question answering systems.  Fig. \ref{fig:sequence} shows the architecture of the sequence to sequence model comprising an encoder and a decoder. Encoder and a decoder are two independent essential neural network models integrated into a massive network to create an output representation sequence. Encoder and decoder comprise two RNNs that act as an encoder and a decoder pair. The encoder receives an input of variable length and maps it to a fixed-length vector, then sends it to the decoder that maps the fixed-length vector back to a variable-length as a target sequence to produce output. This process is accomplished by utilizing the RNN network inside the encoder and decoder.

\subsection{Modular Neural Network }
\begin{figure}
    \centering
    \includegraphics[width=0.4\textwidth]{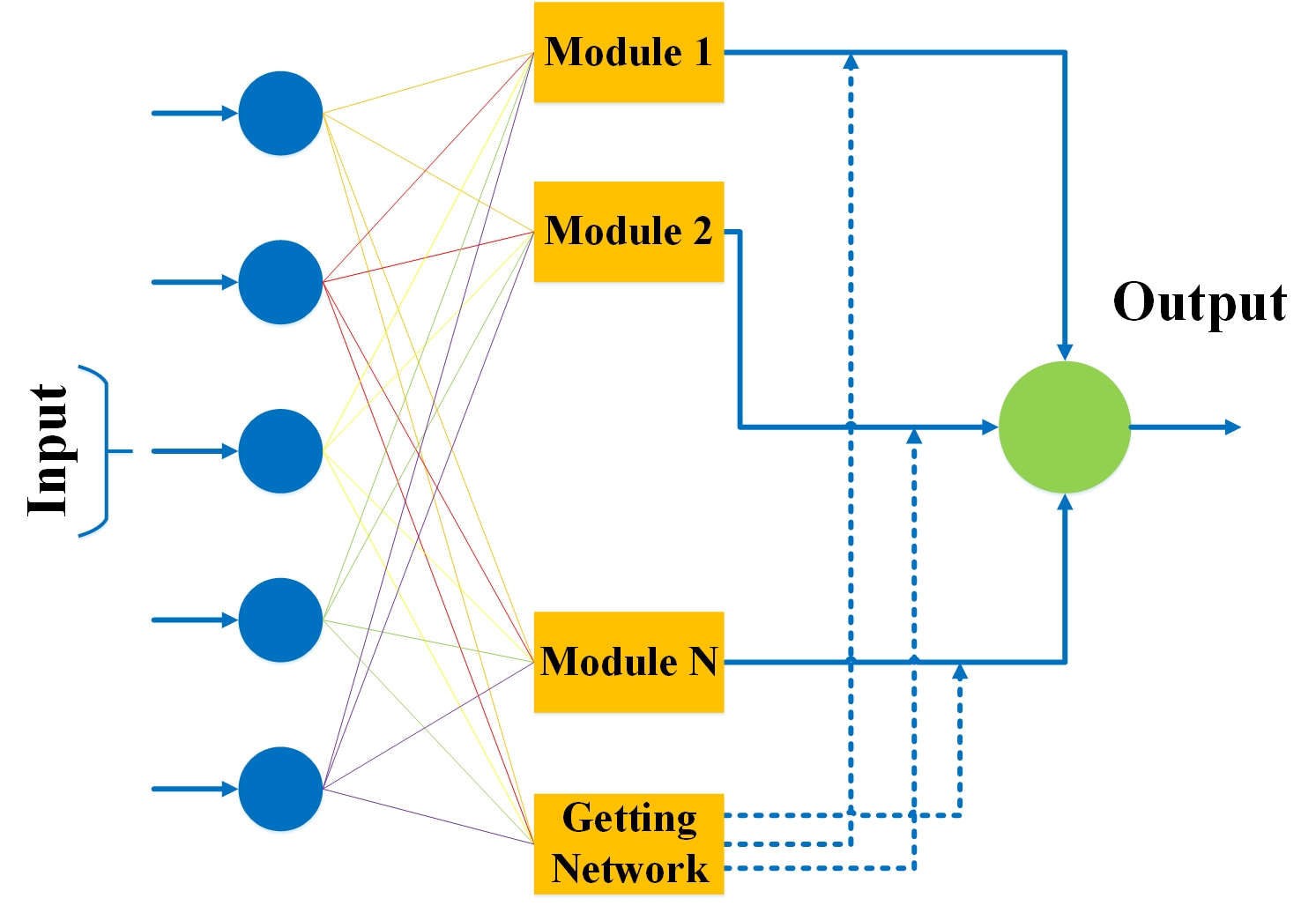}
    \caption{Architecture of modular neural network}
    \label{fig:Modular}
\end{figure}

Last but not least, a modular neural network is a type of neural network that has several dissimilar networks that function self-sufficiently and accomplish sub-tasks \cite{happel1994design}. The different networks do not cooperate with or signal each other during the computation process. They work autonomously to achieve the output, and as a result, a large and complex computational procedure is done expressively faster by contravening it down into self-governing components. The computation speed increases because the networks are not cooperating with or even connected. Modular neural networks are efficient and can perform independent training, but they can move towards target problems \cite{auda1998modular}. Fig. \ref{fig:Modular} shows the architecture of a modular neural network working on the principle of divide and conquer to split large problems down into smaller manageable parts called modules to produce a single output for each using modules network. All the modules are trained independently, and then the output of each module is stored in a new NN model denoted as getting network in Figure \ref{fig:Modular} to produce output.

\section{CNN models for COVID-19 Detection} \label{covidmodels}
In the previous section, we have presented various neural networks. This section highlights the role of CNN in fighting against infectious diseases; especially, COVID-19. Since the COVID-19 outbreak in China in December 2019, it has ruthlessly affected the global population and the global economy. The most crucial part of alleviating the spread of COVID-19 is its timely detection, as it is significantly essential to separate and quarantine the infected patient. The core screening technique used to detect COVID-19 is polymerase chain reaction (PCR) testing; however, it is costly and short in supply. Therefore, an alternative screening method such as radiography examination is used to detect COVID-19, which indicates that people affected with COVID-19 have irregularities in their chest, especially in the lungs. However, a specialist radiologist is required to analyze such radiography images. Therefore, there is an urgent need for an automated system that can analyze radiography images and lessen the workload of radiologists. In this case, neural networks-based X-ray screening is assumed a promising technique to test COVID-19 in asymptomatic patients. The leading interest in developing neural networks-based techniques for COVID-19 detection led to the development of many state-of-the-art neural network models to enhance COVID-19 detection accuracy \cite{abbas2021classification,wieczorek2020neural}. Therefore, we provide a comprehensive review of existing neural network models to thoroughly investigate the working of existing models in terms of their achievements and limitations.

\subsection{Decompose, Transfer, and Compose (DeTraC)}
In \cite{abbas2021classification}, the authors present a deep CNN model (DeTraC)  that indexes COVID-19 chest X-ray images. DeTraC can compact any abnormalities in the image dataset using a class decomposition mechanism. DeTraC model consists of three phases. In the first phase backbone pre-trained CNN model of DeTraC is trained to fetch the deep local features of every image. In the second phase, a sophisticated gradient descent optimization process is used for training. In the last phase, the class composition layer is utilized to improve the final classification of Images. DeTraC model is trained and tested using a combination of two datasets (80 CXR images for normal patients and 116 for COVID-19 patients). The experimental results in \cite{abbas2021classification} show that the DeTraC model exhibits a promising accuracy of 93.1\% (with a sensitivity of 100\%) for identifying COVID-19 X-ray images from normal and severe acute respiratory syndrome cases. However, the small dataset used for training and testing the model degrade its performance in complex scenarios.
	
\subsection{COVID-Net}
Recently, Want et al. used COVID-Net, which is a deep CNN model to classify chest x-ray images into normal and COVID-19 infected patients \cite{wang2020COVID}. To build COVID-Net, the human-driven principled network design prototyping is integrated with machine-driven design consideration to make a personalized network design for recognizing COVID-19 cases from chest radiography images. They have used the COVIDx dataset that comprises 16,576 chest radiography images to train the proposed COVID-Net. COVID-Net was pre-trained on the ImageNet dataset and trained on the COVIDx dataset, attaining accuracy of 92.4\% on the COVIDx test dataset. The accuracy of COVID-Net can further be improved by rearranging the data, utilizing network layers, and increasing the size of the dataset.

\subsection{Coro-Net}
The authors in \cite{khan2020coronet} propose another deep CNN model, CoroNet, to detect COVID-19 infection from chest X-ray images automatically. The foundation of Coro-Net is on Xception architecture, which is pre-trained on ImageNet and then trained on a combined dataset prepared using two publicly available datasets. The dataset used to train Coro-Net comprises 1300 images that achieve an overall accuracy of 89.6\%. Although Coro-Net shows promising results on a small dataset, its efficiency can be increased by using a  large dataset and minimum pre-processing of data.

\subsection{OptCoNet}
The authors in \cite{goel2021optconet} propose an optimized CNN for an automatic diagnosis of COVID-19 (OptCoNet). The OptCoNet building blocks have enhanced feature extraction and classification modules, and it uses Grey Wolf Optimizer (GWO) algorithm to optimize the hyperparameters to train the CNN layers. OptCoNet is trained using the datasets collected from two open-source repositories, comprising a total of 2700 images with 900 COVID-19 infected patients' images and 1800 non-COVID-19 images. The obtained results show that OptCoNet has promising results with an average accuracy of 97.78\%.  Nevertheless, the limited dataset used to train the OptCoNet may lead to the inefficiency of the model.

\subsection{COVID-MTNet}
In \cite{alom2020COVID_mtnet}, the authors present a deep learning neural network model called COVID-MTNet that uses multi-task deep learning to detect COVID-19 from X-ray images. COVID-MTNet uses multiple models for different tasks, such as the classification model for COVID-19 detection and the segmentation model for Region of Interest (ROI) detection. Furthermore, the Recurrent Residual Neural Network (RRCNN) model performs the COVID-19 detection task, and the NABLA-N network executes the infected region segmentation from X-ray and CT images. The dataset used to train the model comprises a total of 5,216 images samples. COVID-MTNet has shown 87.26\% of testing accuracy that indicated good efficiency and reliability. However, data samples may be increased to validate the model's robustness and accuracy.

\subsection{CovNet30}
Another CNN-based model for the automatic diagnosis of COVID-19  from chest X-ray images is presented in \cite{gour2020stacked}.  During the training, several sub-models are obtained from Visual Geometry Group, comprises of 19 layers, 16 convolutional layers, 3 fully connected layers, 5 MaxPooling layers, and 1 SoftMax layer to build a 30-layered CNN model (CovNet30), and the resulting sub-models are arranged together using logistic regression. The CovNet30 model categorizes chest X-ray images into COVID-19, Normal, and Pneumonia classes and uses the COVID19 CXR dataset to train the model.  Their COVID19 CXR dataset consists of 2,764 chest X-ray images collected from three open-source data repositories, and it has achieved an accuracy of 92.74\% for the classification of X-ray images.

\subsection{COVIDPEN}
In \cite{jaiswal2020COVIDpen}, the authors propose a COVID-19 detection technique called COVIDPEN that uses chest X-rays and CT scans. COVIDPEN utilizes a transfer learning technique to identify COVID-19 patients. The dataset used to train the model comprises 746 images samples. As a result, COVINPEN has attained an accuracy of 96\% on the chest X-ray images dataset for the detection of COVID-19. Nevertheless, COVINPEN also shows 96\% of accuracy only for a small dataset.

\subsection{PDCOVIDNet}
Another CNN model, Parallel-Dilated COVIDNet (PDCOVIDNet), is proposed in \cite{chowdhury2020pdCOVIDnet} for COVID-19 identification from chest X-ray images. PDCOVIDNet is trained using the dataset comprising 2,905 chest X-ray images divided into COVID-19 and Normal images. The proposed model has attained average accuracy of 96.58\% that shows the high reliability of the PDCOVIDNet. However, the dataset used to train and test the model is very small, and PDCOVIDNet will likely exhibit degradation in robustness and efficiency for larger datasets.

\subsection{U-Net}
In \cite{kalane2021automatic}, a novel deep learning model is trained on chest X-ray images using U-Net architecture to detect COVID-19. The dataset used to evaluate the model comprises a total of 1000 chest X-ray images. However, it shows poor performance for larger real-time datasets.

\subsection{CapsNet}
The authors in \cite{toraman2020convolutional} present a novel ANN called CapsNet for identifying COVID-19 disease using chest X-ray images with capsule networks. Capsule networks preserve objects' spots and possessions in the image and model their relationships orderly to overcome the pooling layer issues of feature extraction (i.e., missing small features during feature passing to the next layer). Unfortunately, CapsNet also used a small dataset of 1050 images to train the model, and its performance may degrade for a larger dataset.

\begin{table*}[]
\centering
\caption{}
\label{tab:my-table}
\begin{tabular}{|l|l|l|l|}
\hline
\textbf{Reference} &
  \textbf{Model} &
  \textbf{Dataset} &
  \textbf{Accuracy} \\ \hline
\cite{abbas2021classification} &
  \begin{tabular}[c]{@{}l@{}}DeTraC deep convolutional\\ neural network\end{tabular} &
  \begin{tabular}[c]{@{}l@{}}80 samples of normal X-ray \\ images and 105 COVID-19 images\end{tabular} &
  93.1\% \\ \hline
\cite{wang2020COVID} &
  \begin{tabular}[c]{@{}l@{}}COVID-Net deep convolutional\\ neural network\end{tabular} &
  \begin{tabular}[c]{@{}l@{}}76 samples of normal X-ray\\ images and 5526 COVID-19 images\end{tabular} &
  92.4\% \\ \hline
\cite{khan2020coronet} &
  CoroNet deep neural network &
  \begin{tabular}[c]{@{}l@{}}310 samples of normal X-ray\\ images and   330 COVID-19 images\end{tabular} &
  90\% \\ \hline
\cite{goel2021optconet} &
  \begin{tabular}[c]{@{}l@{}}OptCoNet optimized convolutional\\ neural network\end{tabular} &
  \begin{tabular}[c]{@{}l@{}}1,800 samples of normal X-ray \\ images and 900 COVID-19 images\end{tabular} &
  97.78\% \\ \hline
\cite{alom2020COVID_mtnet} &
  \begin{tabular}[c]{@{}l@{}}COVID\_MTNet deep learning \\ model\end{tabular} &
  \begin{tabular}[c]{@{}l@{}}1,341 samples of normal X-ray\\ images and 3,875 COVID-19 images\end{tabular} &
  84.76\% \\ \hline
\cite{jaiswal2020COVIDpen} &
  \begin{tabular}[c]{@{}l@{}}COVIDPEN pruned efficiently \\ net-based model\end{tabular} &
  \begin{tabular}[c]{@{}l@{}}746 total samples of normal X-ray\\ images and COVID-19 images\end{tabular} &
  96\% \\ \hline
\cite{chowdhury2020pdCOVIDnet} &
  \begin{tabular}[c]{@{}l@{}}PDCOVIDNet parallel‑dilated\\ CNN\end{tabular} &
  \begin{tabular}[c]{@{}l@{}}1,341 samples of normal X-ray\\ images and 219 COVID-19 images\end{tabular} &
  96.58\% \\ \hline
\cite{toraman2020convolutional} &
  Capsnet novel artificial neural   network &
  \begin{tabular}[c]{@{}l@{}}1050 samples of normal X-ray\\ images and 231 COVID-19 images\end{tabular} &
  97.24\% \\ \hline
\cite{saha2021emcnet} &
  \begin{tabular}[c]{@{}l@{}}EMCNet CNN\\ and ensemble of machine learning \\ classifiers\end{tabular} &
  \begin{tabular}[c]{@{}l@{}}2,300 samples of normal X-ray\\ images and 2,300 COVID-19 images\end{tabular} &
  98.91\% \\ \hline
\cite{azeemieee} &\begin{tabular}[c]{@{}l@{}}Four layered CNN model for analyzing \\CT images\end{tabular}
   &
  \begin{tabular}[c]{@{}l@{}}3,250 samples of normal ICT\\ images and 5,776 COVID-19 images\end{tabular} &
  97.8\% \\ \hline
\end{tabular}
\end{table*}

\section{ConXNet model for COVID-19 Detection with Large Datasets} \label{ConXNet}
We have discussed various CNN models for COVID-19 detection in the previous section. However, one major issue with those models is that their reliability is not tested for larger datasets. Therefore, this section describes our proposed ConXNet  model that is trained on a large chest X-ray images dataset. In the following, we thoroughly present this novel ConXNet model for COVID-19 detection.

\subsection{Dataset}
As mentioned earlier, we have used a large data set to evaluate the performance of the ConXNet model. The dataset comprises a total of 13,808 chest X-ray images with 3,616 COVID-19 data and 10,192 normal or non-COVID-19 data \cite{web1}. This dataset is acquired from different publically available datasets, online sources, and published papers, such as 2,473 CXR images are collected from the podcast dataset, 183 CXR images from a German medical school, 559 CXR images from SIRM, Github, Kaggle \& Tweeter, and 400 CXR images are collected from another Github source \cite{web2, web3, web4,web5,web6,web7,web8}.

 \subsection{ConXNet Architecture}
\begin{figure*}
    \centering
    \includegraphics[width=0.9\textwidth]{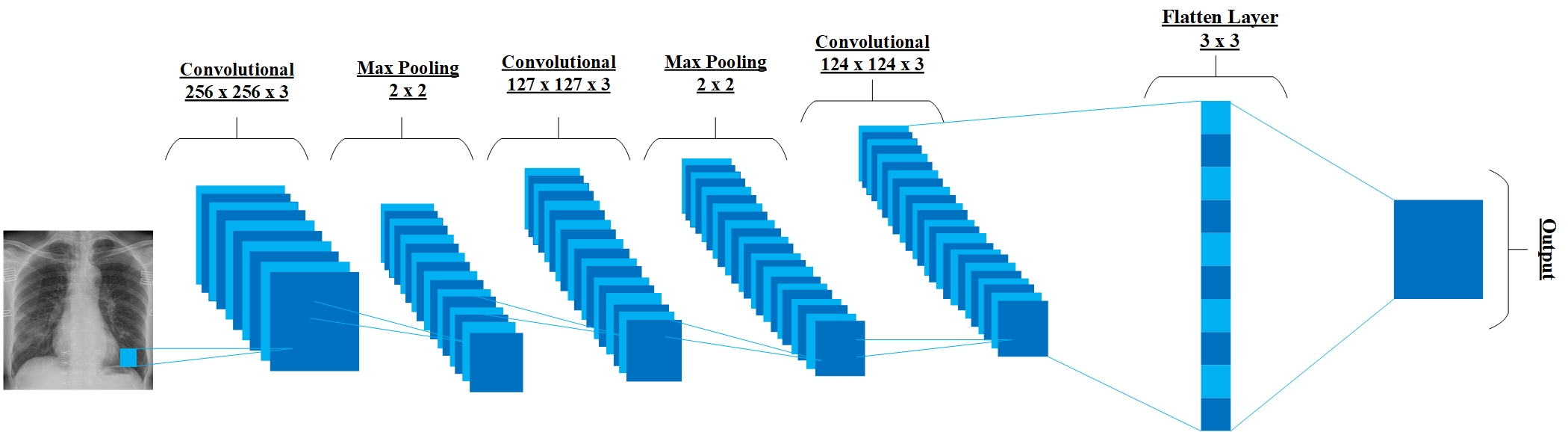}
    \caption{Proposed ConXNet model for COVID-19 detection}
    \label{fig:8}
\end{figure*}

The ConXNet model consists of four blocks where each block comprises a convolutional layer (Conv), rectified linear unit (ReLU) (operating as an activation function), batch normalization, and max-pooling layer. Figure \ref{fig:8} represents the detailed architecture of the proposed ConXNet model. X-ray images are fed to the first Conv layer as an input to extracts the features (edges, soft edges, blur) from the given input. Once the features are extracted, the Conv layer produces a filter matrix called a feature map as an output. After applying different filters on the input image, ReLU is used for non-linear operations so that the model can learn non-negative linear values. Later on, the rectified feature map is passed through the MaxPooling layer to fetch the most prominent element. Furthermore, batch normalization techniques are applied to prevent the model from overfitting. Once all the convolutional operations are performed, the output is flattened before being sent to the fully connected dense layer to produce the final output. The last layer, called the output layer, classifies COVID-19 or Normal X-ray images from our data.

\subsection{Experimental results}
 We have implemented the CNN model in a 64-bit Windows10 operating system using Python 3.6. Tensor flow framework is used to build and train the proposed model using Keras as the backend. Before fitting the model, dataset inconsistencies are removed by selecting equal images from both classes, such as 3,500 for COVID images and 3,500 for Normal images. A total of 7,000 images are used for training and evaluation of the proposed method model. Furthermore, the dataset is split into 70\% (4,900) images for training and 30\% (2,100) images for testing purposes. Also, the model is compiled using binary cross-entropy (BCE) as a loss function for binary classification. To fit the model, epochs size of 100 and batch sizes of 32 are used, while 0.001 is used as a preliminary learning rate value. Finally, Adam optimizer is used to reduce the error rate and automatically tune the learning rate, an efficient variant of gradient descent that prevents the model from the hand-tuning of the learning rate and does it by itself more quickly and efficiently. The results show that this CNN model achieves an overall accuracy of 97.8\%, which is significantly promising for large datasets.  The accuracy is measured as follows:
 \begin{equation}
\label{F1}
\text{Accuracy} = \frac{\text{No.~ of~Images~Correctly~Classified}}{\text{Total~No.~of~Images}}
 \end{equation}




Moreover we have also evaluated the performance of the proposed model by carrying out a real-world evaluation test. In the following, we discuss the results of our test.
\subsubsection{Obtained results for "COVID-19\_Radiography\_Dataset:"}
To effectively observe the performance of the proposed model, the data set is divided into training and testing with the ratio of 70\% and 30\%, respectively. Tensorflow framework is used to build and train the proposed model using Keras as the back end. The model is compiled using binary Crossentropy and accuracy performance metrics. Furthermore, learning is first initialized to 0.001 preliminary to train the model, while 100 epochs are used with 32 batch sizes. Moreover, the error rate is optimized using Adam optimizer for precise learning. Finally, three more common performance measures (accuracy, precision, and F1-Score) are used to observe the proposed model's efficiency. It is observed that the model achieved a significant accuracy score of 97.8\% with the precision and F1-score of 97.93\% and 97.92\%, respectively. Table \ref{tab1} summarizes the results produced by the ConXNet model.
\begin{table}[]
\caption{Performance Measures of ConXNet Model}
\label{tab1}
\begin{tabular}{|l|l|l|l|}
\hline
\textbf{Epochs}           & \textbf{Accuracy}           & \textbf{Precision}           & \textbf{F1-Measure}          \\ \hline
\multicolumn{1}{|c|}{100} & \multicolumn{1}{c|}{97.8\%} & \multicolumn{1}{c|}{97.93\%} & \multicolumn{1}{c|}{97.92\%} \\ \hline
\end{tabular}
\end{table}

\subsubsection{Real-world evaluation}
After training and testing the proposed model on COVID-19\_Radiography\_Dataset, we also carried a real-world evaluation test to evaluate the efficiency and accuracy of the proposed model on unseen data. The real-world evaluation experiment is performed by giving random images as input to classify whether the images belong to COVID positive or negative patients. The chest X-ray image is passed to the network as input, and then the image is preprocessed to match the target size input for the proposed model. Moreover, the proposed model predicts the probabilities of the given image, whether it belongs to the COVID class or normal class, with an approximate predicted percentage. The heat map effect is used to visualize that particular area in the image affected by the disease. These experiments give us a valuable view to evaluate the efficiency of the proposed model in the real world.

\begin{figure*}
     \centering
     \subfloat[]{\includegraphics[width=0.45\textwidth]{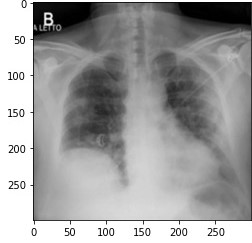}\label{OIC1}} \qquad\qquad
    \subfloat[]{\includegraphics[width=0.42\textwidth]{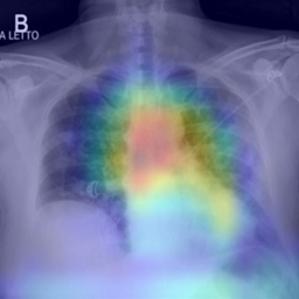}\label{OIC2}} \qquad\qquad
   \newline
      \subfloat[]{\includegraphics[width=0.45\textwidth]{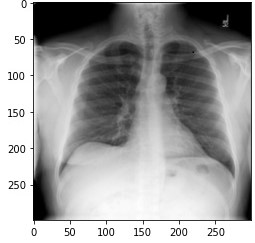}\label{OIC3}} \qquad\qquad
      \subfloat[]{\includegraphics[width=0.42\textwidth]{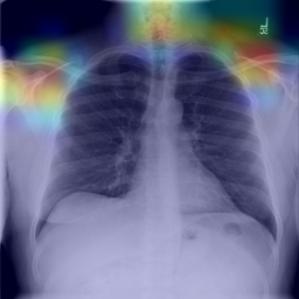}
    \label{OIC4}}
    \caption{Images used to evaluate the model accuracy with their outputs (a) Original image of COVID-19 patient (b) Heat map view of COVID infected region (c) Original image of normal patient (d) Heat map view of normal image}
    \label{Images used to evaluate the model accuracy with their outputs}
\end{figure*}

Figure \ref{OIC1} is illustrates as original X-ray image of COVID patient and Figure \ref{OIC2} illustrates the results after detection of images with heat map highlights of infected regions. While Figure \ref{OIC3} shows the original image of the normal patient and Figure \ref{OIC4} shows after evaluating the results.

\section{CNN models for other infectious diseases detection} \label{diseases}
This section briefly discusses the role of CNN for detection and diagnosis advancements of various other infectious diseases that can improve patient care.

\subsection{Alzheimer's disease detection}
Several studies have investigated the use of CNN for improving Alzheimer's disease detection and diagnosis. For instance, in \cite{vu2018non}, the authors propose an automated and reliable deep learning model for detecting and identifying Alzheimer's disease using magnetic resonance imaging (MRI) and Positron Emission Tomography (PET) that achieved 94.48\% average accuracy. In another work \cite{islam2017ensemble}, an ensemble deep CNN model is developed to detect Alzheimer's disease using the Open Access Series of Imaging Studies (OASIS) dataset that acquired 93.18\% accuracy. In \cite{wen2020convolutional}, a CNN model developed for Alzheimer's disease detection use a brain imaging structured dataset comprising brain MRI images that achieved 96\% accuracy, indicating an excellent efficiency as compared to other state-of-the-art models. In \cite{farooq2017deep}, a 4-way CNN-based classifier is used that attain 98.8\% accuracy by classifying four classes, including Alzheimer's disease, mild cognitive impairment, late mild cognitive impairment, and no disease using the ADNI dataset. Recently, Alzheimer's disease is detected by introducing transfer learning from a dataset of 2D images to 3D CNNs to enable early diagnosis of Alzheimer's disease using MRI imaging datasets \cite{ebrahimi2021convolutional}.

\subsection{Cancer detection}
Deep learning methods, such as CNN that can extract hierarchical features from image data without manual selection, are successfully applied in cancer tissue detection. For example, early breast cancer detection, diagnosis, and treatment are possible due to a Computer-Aided Diagnostic (CAD) system based on mammograms analyzed by CNN-based classifiers. In \cite{wang2019breast}, a CAD approach based on deep features detects breast cancer using mammograms images dataset, achieving 81.75\% accuracy. In \cite{alakwaa2017lung}, a novel CNN-based model detects lung cancer using Computer Tomography (CT) scan images, and achieved 86.6\% accuracy. In \cite{yoo2019prostate}, a novel CNN-based model is proposed for detecting prostate cancer using diffusion-weighted magnetic resonance imaging (DWI) dataset, which acquired 84\% accuracy. In \cite{ben2016fully}, a Fully Convolutional Network (FCN) is introduced to detect liver cancer using CT scan images, achieving 86\% accuracy by utilizing a 3-fold cross-validation technique.

Besides breast cancer detection, CNN has also shown a great potential for brain tumor detection, the most frequent and severe type of cancer with a life expectancy of only a few months in the most advanced stages. Therefore, treatment planning in brain tumors is an essential step in improving patients' quality of life. In \cite{abiwinanda2019brain}, a CNN classification model is proposed for automatic brain tumor detection using brain MRI images datasets. The experimental results show that the proposed model acquired 97.5\% accuracy. Another CNN model presented in \cite{seetha2018brain} classifies images for brain tumor detection and achieves 97.87\% accuracy. The CNN model presented in \cite{kaldera2019brain} performs the classification and segmentation of brain tumors using the MRI images dataset and achieves 94.6\% accuracy. In \cite{sarhan2020brain}, a CNN-based model segments brain tumors from the MRI images dataset, and the model acquires 96\% of average accuracy. Recently, reference \cite{kalaiselvi2020deriving} introduced a CNN model for classifying benign tumors that trains and test the model using the BraTS2013 and WBA dataset, respectively, and attains 96-99\% accuracy.

\subsection{Retinal disease detection}
Optical Coherence Tomography (OCT) imaging is used to make accurate and reliable diagnoses of retinal disorders, which is considered critical for clinical relevance. Rajagopalan et.al. in \cite{rajagopalan2020deep} introduced a novel CNN model for accurate detection and classification into normal, drusen macular degeneration and diabetic macular edema using Optical Coherence Tomography (OCT) imaging dataset. Moreover, to reduce intrinsic speckle noise, the Kuan filter is used to input OCT pictures first. Furthermore, hyperparameters optimization techniques are used to optimize the CNN network, and K-fold validation is performed to assure complete utilization of the dataset. As a result, the presented model achieved 95.7\% of accuracy. Another CNN model presented in \cite{soomro2019impact} accurately detects retinal blood vessels.
Recently, in \cite{akil2021detection}, the CNN-based approach localizes, identifies, and quantifies abnormal features in the eye retina using OCT images dataset and achieves an accuracy of 95.8\%. In \cite{tayal2021dl}, OCT imaging of the retinas is analyzed using three different CNN models comprising five, seven, and nine layers to identify the various retinal layers, extract usable information, detect new aberrations, and predict several eye abnormalities.  Furthermore, the CNN model recently developed in \cite{mittal2021retinal} classifies retinal disease using OCT images and achieves 98.73\% accuracy.

\begin{table*}
\centering
\arrayrulecolor{black}
\begin{tabular}{|l!{\color{black}\vrule}l|l|l|l|}
\arrayrulecolor{black}\hline
\textbf{ Diseases }        & \textbf{ Ref. } & \textbf{ Model }                                                                                                                                    & \textbf{ Dataset }                                                                                                              & \textbf{ Accuracy }  \\
\hline
\multirow{4}{*}{Alzheimer} & \cite{vu2018non}              & \begin{tabular}[c]{@{}l@{}}Deep CNN for Alzheimer’s\\disease detection\end{tabular}                                                                 & Total 615 MRI scans images                                                                                                      & 94.48\%              \\
\arrayrulecolor{black}\cline{2-5}
                           & \cite{islam2017ensemble}              & An ensemble of deep CNN                                                                                                                             & \begin{tabular}[c]{@{}l@{}}OASIS dataset comprises of \\total 416 MRI scans images\end{tabular}                                 & 93.18\%              \\
\cline{2-5}
                           & \cite{wen2020convolutional}              & \begin{tabular}[c]{@{}l@{}}CNN for Classification of \\Alzheimer’s disease\end{tabular}                                                             & \begin{tabular}[c]{@{}l@{}}ADNI dataset comprises of total \\1455 MRI scans images\end{tabular}                                 & 96\%                 \\
\cline{2-5}
                           & \cite{farooq2017deep}              & \begin{tabular}[c]{@{}l@{}}A Deep CNN based multi-class \\classification of Alzheimer’s disease\end{tabular}                                        & \begin{tabular}[c]{@{}l@{}}Dataset comprises of total 355 \\MRI scans images\end{tabular}                                       & 98.8\%               \\
\arrayrulecolor{black}\hline
\multirow{9}{*}{Cancer}    & \cite{wang2019breast}              & \begin{tabular}[c]{@{}l@{}}Breast cancer Detection using \\extreme learning machine based \\on feature fusion with CNN deep \\features\end{tabular} & Total 400 Mammography images                                                                                                    & 81.75\%              \\
\arrayrulecolor{black}\cline{2-5}
                           & \cite{alakwaa2017lung}              & \begin{tabular}[c]{@{}l@{}}Lung cancer detection and class-\\ification with 3D CNN\end{tabular}                                                     & \begin{tabular}[c]{@{}l@{}}Kaggle Data Science Bowl (DSB) \\comprises of total 1397 MRI \\scans images\end{tabular}             & 86.6\%               \\
\cline{2-5}
                           & \cite{yoo2019prostate}              & \begin{tabular}[c]{@{}l@{}} Prostate cancer detection using \\deep CNN\end{tabular}                                                                  & \begin{tabular}[c]{@{}l@{}}Diffusion-weighted magnetic \\resonance Imaging dataset \\comprises of total 427 images\end{tabular} & 84\%                 \\
\cline{2-5}
                           & \cite{ben2016fully}              & \begin{tabular}[c]{@{}l@{}}Fully CNN for liver segmentation and\\lesions detection\end{tabular}                                                     & \begin{tabular}[c]{@{}l@{}}Dataset comprises of total 88 \\CT scan images\end{tabular}                                          & 86\%                 \\
\cline{2-5}
                           & \cite{abiwinanda2019brain}              & Brain tumor classification using CNN                                                                                                                & \begin{tabular}[c]{@{}l@{}}Radiopaedia and brain tumor \\image segmentation benchmark\\2015 datasets are used\end{tabular}      & 97.5\%               \\
\cline{2-5}
                           & \cite{seetha2018brain}              & Brain tumor detection using CNN                                                                                                                     & \begin{tabular}[c]{@{}l@{}}BRATS dataset comprises of 217 \\MRI images\end{tabular}                                             & 97.87\%              \\
\cline{2-5}
                           & \cite{kaldera2019brain}              & \begin{tabular}[c]{@{}l@{}}Brain tumor classification and segmen-\\tation using faster R-CNN\end{tabular}                                           & Total 218 MRI images are used                                                                                                   & 94.6\%               \\
\cline{2-5}
                           & \cite{sarhan2020brain}              & \begin{tabular}[c]{@{}l@{}}Brain tumor classification in magnetic \\resonance images using deep learning \\and wavelet transform\end{tabular}       & MRI images dataset                                                                                                              & 96\%                 \\
\cline{2-5}
                           & \cite{kalaiselvi2020deriving}              & \begin{tabular}[c]{@{}l@{}}Deriving tumor detection models using \\CNN from MRI of human brain scans ~\end{tabular}                                 & \begin{tabular}[c]{@{}l@{}}BraTS2013 dataset and WBA \\dataset are used\end{tabular}                                            & 96-99\%              \\
\arrayrulecolor{black}\hline
\multirow{4}{*}{Retinal}   & \cite{rajagopalan2020deep}              & \begin{tabular}[c]{@{}l@{}}Deep CNN framework for retinal disease \\diagnosis\end{tabular}                                                          & \begin{tabular}[c]{@{}l@{}}OCT images dataset containing a\\total of 12,000 images\end{tabular}                                 & 95.7\%               \\
\arrayrulecolor{black}\cline{2-5}
                           & \cite{akil2021detection}              & \begin{tabular}[c]{@{}l@{}}Detection of retinal abnormalities Using\\~CNN\end{tabular}                                                              & \begin{tabular}[c]{@{}l@{}}Dataset comprise of a total 1110 \\fundus images\end{tabular}                                        & 95.8\%
  ~           \\
\cline{2-5}
                           & \cite{tayal2021dl}              & \begin{tabular}[c]{@{}l@{}}DL‑CNN‑based approach diagnosis of \\retinal diseases\end{tabular}                                                       & \begin{tabular}[c]{@{}l@{}}OCT images dataset comprise of \\a total 84,495 images\end{tabular}                                  & 96.5\%               \\
\cline{2-5}
                           & \cite{mittal2021retinal}              & \begin{tabular}[c]{@{}l@{}}Retinal disease classification using CNN \\algorithm\end{tabular}                                                        & \begin{tabular}[c]{@{}l@{}}OCT images dataset comprise of \\a total 108,312 images\end{tabular}                                 & 98.73\%              \\
\arrayrulecolor{black}\hline
\end{tabular}
\end{table*}

\section{Future Research Directions} \label{directions}
In this section, we present future research directions and challenges such as complexity of the algorithms, insufficient available data, privacy and security, and integration of biosensing with ANNs. These research directions require considerable attention for improving the scope of ANNs for medical diagnostic and treatment
\subsection{Complexity}
The complexity of neural network-based models for disease detection and diagnosis depends on the time and number of samples used to train the network. Since in medical imagining applications, the accurate detection of disease is more important, thus the analysis complexity (such as time and computational complexities) is compromised in most of the cases \cite{mikolov2011extensions,ludermir2006optimization}. However, in the future, neural network-based disease detection applications can introduce different methods to address the complexity issue for preventing excessive resource consumption without degrading accuracy. For example, ID and 2D filters can be used at the feature-extraction image-restoration layer with high-resolution images to reduce the complexity \cite{park2020low}. Furthermore, data transformation techniques including data augmentation \cite{koning2019reducing}, variable standardization \cite{shanker1996effect}, and suitable initialization of ANN \cite{bowden2003data} can also play a significant role in reducing complexity since these techniques use limited data samples to achieve better results without compromising on accuracy. Moreover, limiting the number of neurons during model training \cite{labach2019survey}, dropout layer technique \cite{srivastava2013improving}, network pruning techniques \cite{liu2020pruning}, normalization techniques \cite{srinidhi2020deep}, transfer learning techniques, and brute-force exploration approach \cite{robinson2018brute} can also be used to minimize complexity.  Nevertheless, the research on developing low complexity neural networks for infectious diseases detection is still in infancy and needs the researchers' attention.

\subsection{Algorithm selection}
Another challenging issue for neural networks-based infectious diseases detection is the selection of the proper algorithm. In ANNs, algorithms are divided into two main categories such as supervised learning and unsupervised learning. ANNs based models comprises numerous algorithms such as decision trees (DT) \cite{singh2014comparative}, support vector machine (SVM) \cite{chandra2018survey}, Naïve Bayes (NB) \cite{kulkarnirandom}, k-means \cite{blomer2016theoretical}, k-nearest neighbor (KNN) \cite{bhatia2010survey}, and random forest (RF) \cite{saiyed2016survey} for completing different tasks. In addition, algorithm selection according to the considered dataset is also necessary as inappropriate algorithm selection fails to provide the required results. Supervised learning deals with labeled data and has both the input and output features to map the targets that can be very useful for medical image analysis. However, unsupervised learning deals with unlabeled and uncategorized data to find hidden structures. Therefore, if classes have no training data or missing values, then there is a probability that this technique lacks classification data efficiency. Consequently, unsupervised learning is performed on unseen data. Hence, supervised learning is preferred over unsupervised learning, particularly for image classification tasks.
The suitable algorithms should have several attributes such as validity (formative the accuracy of an algorithm for data clustering), stability (the difference of results attained in diverse implementations must be related to each other), and scalability (the volume of clustering of large capacities of data in an efficient way) \cite{ruiz2017survey}. Therefore, further investigation is required to find the optimal neural network algorithm for a specific disease where one neural network may perform well for detecting a particular disease but may fail in the other case.
\subsection{Deficient Training Data}
 In addition to algorithm selection, another major issue is getting valuable information from the raw data. The ANNs are mostly applied on large data sets and composite models that need extensive training to obtain the required training effects. However, acquiring adequate training data is challenging due to large data unavailability in numerous domains such as diseases detection using image classification. Thus, the lack of available data for training and evaluating the parameters in the neural networks results in network inefficiency and overfitting. For instance, blood pressure measurements in most of the COVID-19 critical patients are erroneous and unstable, which limits the amount of accurate data.
 In existing literature \cite{pomerleau1991efficient}, various methods such as dropout \cite{hou2019weighted} and data augmentation \cite{ding2016convolutional} are introduced to decrease the change. Although there is a large amount of data available in several other domains where big data analytics is performed, however, there is a deficiency of available data in medical imaging analysis and infectious diseases detection, which is quite challenging for the ANNs.

\subsection{Privacy and Security}
One of the key issues with using ANNs for infectious diseases detection is privacy and security concerns. Nevertheless, reference \cite{barni2006privacy} introduced several methods to attain privacy from the “trusted” authority, attackers, and involved entities. Recently, for COVID-19 location-aware apps, a polling mechanism is used to ensure the privacy of the patient infected with COVID-19 from the non-infected patient \cite{chowdhury2020COVID}. This allows non-infected patients to poll the health authority frequently to check whether they have been in close contact with infected patients. Moreover, privacy in the healthcare domain can also be achieved using private messaging systems \cite{tyagi2017stadium}, private set intersection protocols \cite{de2010practical}, and memo verification methods using cryptography \cite{zheng2020industrial}. However, these protocols are computationally intensive, requiring a trade-off between security and computational efficiency \cite{althunibat2013trade}. Therefore, in the future low computational techniques such as blockchain \cite{goel2019deepring}, cloud computing \cite{kwabena2019mscryptonet}, and Deep Belief Neural Networks (DBNN) \cite{manimurugan2020effective} require more focus to ensure the required level of privacy and security.

\subsection{Integration of Bio-sensing and ANN}
The tremendous progress in nanotechnology over the past few years has also advanced the development of biosensors for medical applications \cite{javaid2021temperature, saeed2020body}. Biosensors are analytical devices that can detect particular biological strains and provide cellular level monitoring at the nanoscale \cite{sin2014advances,fahim2020efficient}. The research communities are focusing on integrating AI intelligence methods such as metaheuristic algorithms, and ANN models with biosensing applications to overcome their challenges for intelligent data detection in a biological medium for infectious disease detection \cite{javaid2020feedforward,fahim2019fuzzy}. Moreover, ANN’s adoption in biosensing applications can significantly improve disease detection accuracy and reliability as biosensors can produce data in time-series form that can be trained and tested using state-of-the-art NNs \cite{cui2020advancing}. For example, in \cite{ferentinos2012use}, biosensors generated time-series data to detect pesticides with 85\% success rate. In another work, \cite{maleki2017novel}, a novel biosensor and ANN-based integrated system detected catechol from water samples with 99.7\% accuracy.  Nevertheless, the integration of biosensing and ANN for medical application is in the initial phase; more dedicated efforts are required to integrate these two technologies to achieve great benefits in the medical domain.

\section{Conclusion} \label{conclusion}
In this paper, we reviewed thoroughly the advances in infectious disease detection and diagnostics brought by the ANNs. The detailed discussions reveal that ANN’s ability to model the human brain to solve complex problems has a huge tendency to improve disease detection and diagnostic accuracy, leading to advanced patient care and treatment. In addition to the review of the existing ANN-based disease detection models, we have also proposed a new model ConXNet that considerably enhances the detection accuracy of COVID-19 patients.  The extensive testing and training of ConXNet using different available datasets show that the ConXNet can detect COVID-19 patients more accurately than the existing models. Finally, we presented several future directions to emphasize the gap. Since this work is mainly focused on the ANN-based disease detection models, it will be interesting to investigate the potential of ANN for targeted drug delivery and advanced surgical procedure in the future. Moreover, we also suggest testing different evolving techniques such as transfer learning and hybrid schemes to enhance detection accuracy and precision in medical diagnosis.

\begin{IEEEbiography}[{\includegraphics[width=1in, height=1.25in, clip, keepaspectratio]{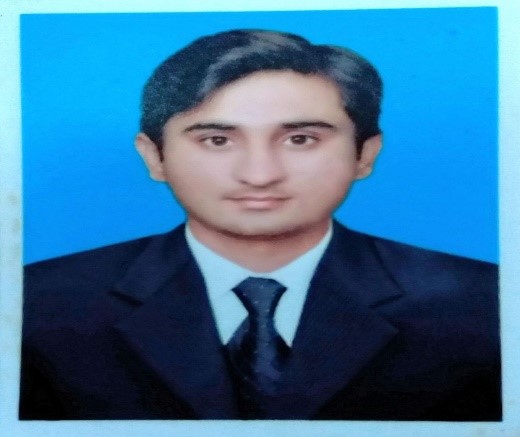}}]
{Muhammad Azeem} received his B. S. degree in Information Technology from the University of Sargodha (UoS), Pakistan, in 2016. Currently, he is pursuing his M. S. degree in Computer Science from the University of Okara (U.O.), Pakistan. His research interests include machine learning, deep learning, and wireless network communication.
\end{IEEEbiography}

\begin{IEEEbiography}[{\includegraphics[width=1in, height=1.25in, clip, keepaspectratio]{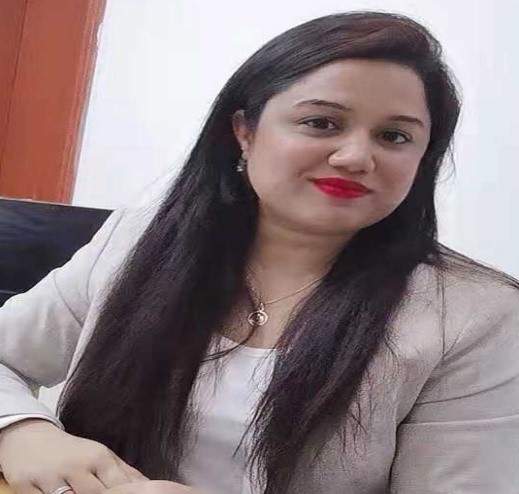}}]
{SHUMAILA JAVAID} did her Ph.D. in Computer Science from Shaanxi Normal University, China. She is working as an Assistant Professor at Ilma University Karachi, Pakistan, and she is also associated with Tongji University, Shanghai, China as a Postdoctoral researcher. She received her M.S. degree in Telecommunication and Networking from Bahria University, Pakistan, in 2015. She completed her B.S. degree from COMSATS University (C.U.), Pakistan, in 2012. Her main research interests include Wireless Sensor, Intrabody Nanonetworks, Robotic Communication, Neural Networks,  Information-Centric Networks, and Wireless Networking in general.
\end{IEEEbiography}

\begin{IEEEbiography}[{\includegraphics[width=1in, height=1.25in, clip, keepaspectratio]{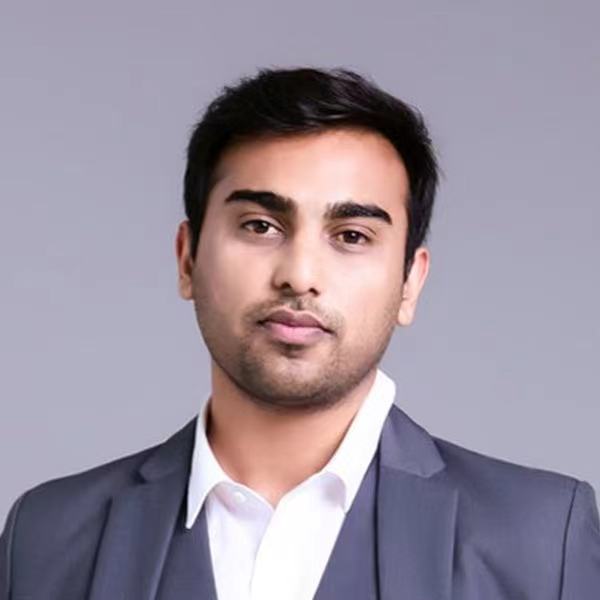}}]
{Hamza Fahim} is associated with Tongji University, Shanghai, China as a Postdoctoral researcher. He did his Ph.D. in Computer Science and Networks from Xi’an Jiaotong University, China. He is also associated with Ilma University Karachi, Pakistan, as Assistant Professor. He holds his B.S. and M.S. degrees from the computer science department, COMSATS University (C.U.), Pakistan, in 2012 and 2015. His current research interests include Intrabody Nanonetworks, Software Defined Networks, and Wireless Sensor Networks.
\end{IEEEbiography}

\begin{IEEEbiography}[{\includegraphics[width=1in, height=1.25in, clip, keepaspectratio]{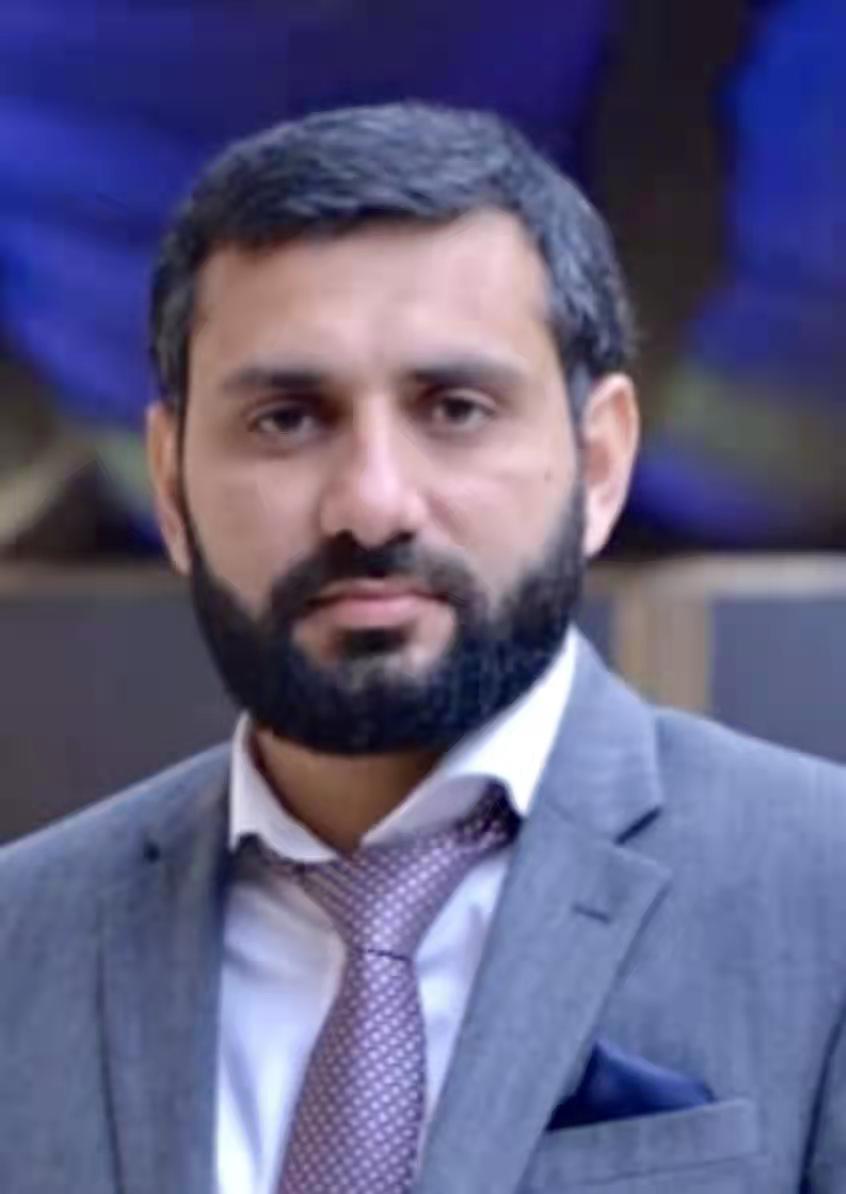}}]
{NASIR SAEED} (Senior Member, IEEE) received the bachelor's degree in telecommunication from the University of Engineering and Technology, Peshawar, Pakistan, in 2009, the master's degree in satellite navigation from the Polito di Torino, Italy, in 2012, and the Ph.D. degree in electronics and communication engineering from Hanyang University, Seoul, South Korea, in 2015. He was an Assistant Professor with the Department of Electrical Engineering, Gandhara Institute of Science and IT, Peshawar, from August 2015 to September 2016. He has worked as an Assistant Professor with IQRA National University, Peshawar, from October 2016 to July 2017. He is currently a Postdoctoral Research Fellow with the Communication Theory Laboratory, King Abdullah University of Science and Technology (KAUST). His current research interests include cognitive radio networks, underwater wireless communications, aerial networks, dimensionality reduction, and localization.
\end{IEEEbiography}

\bibliographystyle{IEEEtran}
\bibliography{arXiv}

\end{document}